\documentclass[runningheads]{llncs}

\usepackage{eccv}

\usepackage{eccvabbrv}

\usepackage{graphicx}
\usepackage{booktabs}
\usepackage{soul}
\usepackage{tabularx}
\usepackage{multirow}
\usepackage{wrapfig}
\usepackage{makecell}
\usepackage{amsmath}
\newcolumntype{M}[1]{>{\centering\arraybackslash}m{#1}}
\newcommand{\metricw}{1.5cm}

\DeclareMathOperator*{\argmin}{arg\,min}
\usepackage{ulem}
\usepackage{marvosym}

\usepackage[accsupp]{axessibility}  

\usepackage{hyperref}

\usepackage{orcidlink}

\makeatletter
\newcommand{\printfnsymbol}[1]{%
  \textsuperscript{\@fnsymbol{#1}}%
}
\makeatother

\begin{document}

\title{Training-free Controllable Human Motion Generation under Heterogeneous Constraints} 

\titlerunning{Controllable Motion Generation under Heterogeneous Constraints}

\author{Xiaofei Hui\inst{1}\orcidlink{0000-0002-9258-5768} \and
Bo Yan\inst{2}\orcidlink{0009-0003-7483-4032} \and
Haoxuan Qu\inst{1}\orcidlink{0000-0001-5054-3394} \and
Hossein Rahmani\inst{1}\orcidlink{0000-0003-1920-0371} \and
Jun Liu\inst{1}\textsuperscript{\Letter}\orcidlink{0000-0002-4365-4165}
}

\authorrunning{X.~Hui et al.}

\institute{Lancaster University, United Kingdom \and Shenzhen International Graduate School, Tsinghua University, China
\email{\{x.hui,h.qu5,h.rahmani,j.liu81\}@lancaster.ac.uk, yanb22thu@gmail.com}}

\maketitle
\def\thefootnote{{\textsuperscript{\Letter}}}\footnotetext{Corresponding Author}

\begin{abstract}
    Training-free controllable motion generation has attracted growing interest for enabling flexible constraint enforcement without constraint-specific training.
    However, existing training-free methods require constraints to be \textit{continuous objective-based} with differentiable losses, while many real-world requirements are \textit{criterion-based} and provide only discontinuous, sparse, or even black-box feedback. In this paper, we propose Motion-Inference-as-Control (MIC), the first training-free motion generation framework that handles both continuous objective-based and criterion-based motion constraints under a shared mechanism. The key idea is to cast diffusion-based motion generation as a stochastic control problem. This perspective not only provides principled and practically effective step-wise control laws that support criterion-based constraints without requiring differentiability and naturally accommodate objective-based constraints as a special case, but also motivates a control-oriented constraint coordination mechanism that adaptively balances and reconciles motion constraints during generation. Experiments across diverse constraint settings demonstrate the effectiveness of our framework.
\end{abstract}

\section{Introduction}
\label{sec:intro}

Generating human motion is a fundamental problem in computer vision \cite{10313063,Guo_2022_CVPR,tevet2023human,yuan2023physdiff}, with broad applications in virtual reality \cite{8695841,9714119}, game character animation \cite{10.1145/3272127.3275071}, and humanoid robot control \cite{cheng2024expressive}. These applications often require motions to satisfy diverse constraints (e.g., environmental restrictions, user controls, and motion feasibility). A common paradigm to achieve this is to apply constraint-specific model training or adaptation \cite{xie2024omnicontrol,Karunratanakul_2023_ICCV,pinyoanuntapong2025maskcontrol,yuan2023physdiff,li2025morph,tevet2025closd,tan2025sopo}. Yet, as constraint settings can vary substantially across scenes and tasks, such a paradigm may require retraining or adaptation when settings change, 
which limits scalability.

\begin{figure}
    \centering
    \includegraphics[width=0.6\linewidth]{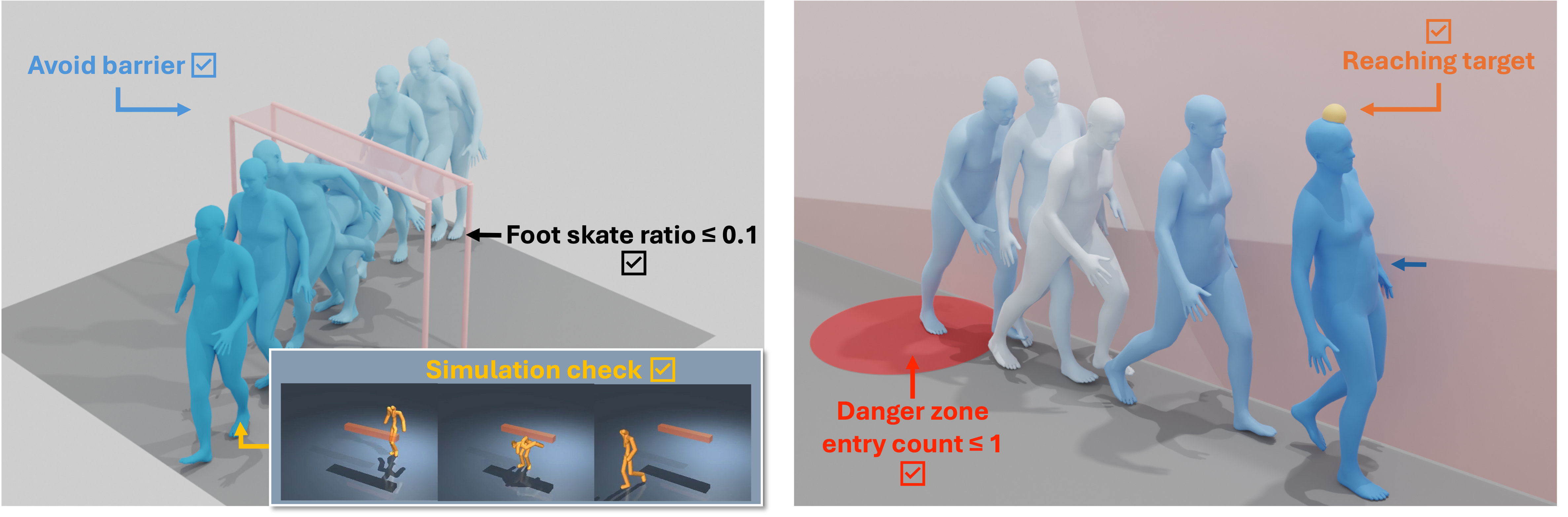}
    \caption{Real-world motion constraints can be heterogeneous. Some are \textit{continuous objective-based} constraints that can be naturally modeled with differentiable losses (e.g., reaching a target point), whereas many others are \textit{criterion-based} and provide feedback only as sparse, event-triggered, or binary evaluations, such as simulation-based validity tests and safety-threshold rules (e.g., entering the danger zone at most once). In practice, these heterogeneous constraints may be required simultaneously within a single motion. 
    The darker mesh colors indicate later frames in the sequence.
    }
    \label{fig:intro}
\end{figure}

Recently, training-free controllable motion generation that bypasses constraint-specific model training has attracted growing attention \cite{liu2024programmable,karunratanakul2024optimizing}. Specifically, as diffusion-based motion generation models \cite{tevet2023human,wen2025hy,zhang2024motiondiffuse} offer both high-quality motion generation and substantial inference-time flexibility, recent works \cite{shafir2023human,liu2024programmable} have explored enforcing constraints via inference-time modulation on diffusion models, often leveraging gradients computed from differentiable constraints.
Yet, despite the promising advances, challenges still remain in effectively integrating diverse real-world constraints in such a training-free manner. Due to the broad range of downstream applications and tasks, motion constraints can be intrinsically heterogeneous and complex, as illustrated in \cref{fig:intro}.   
Some constraints are \textit{continuous objective-based}, naturally expressed as differentiable losses (e.g., reaching target point or maintaining motion velocity), while many constraints are \textit{criterion-based}, providing feedback mainly via event-triggered rules or external evaluators (e.g., safety thresholds, contact events, or feasibility checks) instead of a smooth objective. \textbf{This heterogeneity poses the following key challenges.} (1) Criterion-based constraint signals are often discontinuous, sparse, or even black-box (accessible only via query-based evaluation without analytic gradients), making them difficult to incorporate into existing training-free motion generation pipelines that generally rely on differentiable objectives \cite{liu2024programmable,shafir2023human}. 
This calls for an inference-time mechanism capable of handling criterion-based constraints effectively without requiring differentiability. Moreover, as both constraint types can co-exist in practice, this mechanism would ideally also accommodate continuous objective-based constraints under a unified framework, eliminating the need for separate pipelines and facilitating coherent enforcement of heterogeneous constraints.
(2) Even with such a mechanism, a further challenge arises when multiple heterogeneous constraints are imposed on the same motion. In practice, these constraints may differ substantially in signal strength and spatial-temporal scope (e.g., full sequence or short time windows, full body or local joints). 
Naive combination of these heterogeneous signals can lead to \textit{constraint imbalance}, where stronger signals may dominate weaker ones, and \textit{scope interference}, where local constraints may interfere with global ones.

In this paper, we aim to tackle the above challenges and develop a training-free, plug-and-play framework that progressively steers the motion diffusion process to satisfy heterogeneous constraints simultaneously and coherently. Essentially, this can be viewed as a step-wise decision problem under motion diffusion dynamics, which decides at each denoising step, how to incorporate constraints into the generation process such that the final generated motion satisfies the constraints.
This problem naturally connects to stochastic control~\cite{pavon1989stochastic}, which also seeks to determine, at each timestep, how to drive system dynamics toward a desired outcome. 
Grounded in this connection and inspired by \cite{huang2024symbolic,Kim_2025_ICCV,dou2026constrained,Zhang_2026_CVPR}, we develop Motion-Inference-as-Control (MIC), a principled and practical framework that formulates inference-time constraint incorporation in diffusion-based motion generation as a stochastic optimal control problem. Notably, this control perspective offers key advantages. Specifically, from stochastic optimal control theory, the desired step-wise diffusion guidance for controllable motion generation can be characterized under a control formulation. For criterion-based constraints, the resulting optimal control law can be expressed using only forward constraint evaluations (e.g., binary feedback and evaluator-dependent feasibility scores), eliminating the dependence on gradient computation. Meanwhile, the same framework naturally encompasses continuous objective-based constraints as a special case, where the optimal control law can also be expressed in the form of gradients. Essentially, the above control-theoretic view provides a unifying principle that integrates heterogeneous constraints into a shared control interface: criterion-based and continuous objective-based constraints alike can be modeled as control signals to steer the motion generation, which directly addresses the first key challenge.

While the above formulation allows heterogeneous motion constraints to be cast as control signals in a unified control interface, naively aggregating multiple such signals can easily lead to constraint imbalance and scope interference, undermining constraint satisfaction and motion coherence. We note that the control-system structure of the above formulation turns this coordination challenge into a more structured design problem. Particularly, it exposes two complementary design axes: how much step-wise control effort to assign to each constraint over time, and how to distribute the resulting effects across joints and frames. Building on this, we propose a bilevel constraint coordination mechanism with two components: a \textit{feedback regulator} that monitors constraint satisfaction and adaptively adjusts the relative importance of each constraint throughout the denoising process; and a \textit{control allocator} that distributes the limited step-wise control budget across constraints according to their effective scopes, seeking to best satisfy all requirements while minimizing inter-constraint interference. Together, this bilevel design mitigates control imbalance and scope interference, enabling stable constraint enforcement while preserving motion naturalness.

Overall, our contributions are: (1) By adopting a control-theoretic perspective, we construct a unified control interface for heterogeneous constraint enforcement in motion generation. To the best of our knowledge, this is the first training-free framework for controllable diffusion-based motion generation that enables both criterion-based and continuous objective-based constraints to be jointly enforced within a single inference-time control formulation.
(2) Based on this foundation, we further propose a practical framework that coordinates different constraint control signals via a constraint coordination mechanism, mitigating control imbalance and scope interference while preserving motion naturalness. (3) Experiment results show that our method consistently achieves strong performance with different constraint settings, demonstrating its effectiveness.

\section{Related Work}
\label{sec:related_work}

\textbf{Text-to-Motion Generation.}
Text-to-motion generation aims to synthesize realistic human motion sequences conditioned on text descriptions \cite{10313063,tevet2023human,zhang2024motiondiffuse,jiang2023motiongpt}. To tackle this problem, various methods have been proposed, including variational autoencoders (VAEs) \cite{Guo_2022_CVPR,Zhong_2023_ICCV,cai2021unified}, generative adversarial networks (GANs) \cite{wang2020learning,degardin2022generative}, generative masked modeling methods \cite{guo2024momask,jeong2025hgm}, and auto-regressive-based methods \cite{jiang2023motiongpt,zhang2023generating}.
Amid the broad success of diffusion models in generative modeling \cite{10.1145/3626235,xu20246d,li2026diffgraph,Li_2025_CVPR,gong2023diffpose,foo2023distribution}, diffusion-based methods \cite{tevet2023human,zhang2024motiondiffuse,wen2025hy,chen2023executing} have emerged as a mainstream paradigm for text-to-motion due to their strong ability to generate temporally coherent and diverse motions, motivating a growing body of research building on such diffusion-based models to improve motion generation via both training-based pipelines \cite{zhou2024emdm,bae2025less,yuan2023physdiff,tevet2025closd,li2025simmotionedit,han2025reindiffuse,girolamo2026no} and training-free mechanisms \cite{chen2024pay,zhuo2025infinidreamer,yu2025plug,NEURIPS2024_a28af221,shen2024understanding,ota2025pino}. In this paper, we focus on leveraging diffusion-based models to improve controllable motion generation in a training-free manner.

\textbf{Controllable and Constraint-aware Motion Generation.}
Recent research has explored controllable motion generation that incorporates additional constraint requirements, such as target trajectories, keyframe constraints, and body-part goals \cite{liu2024programmable,Karunratanakul_2023_ICCV,pinyoanuntapong2025maskcontrol,guo2025motionlab}. Most existing approaches achieve controllability through task-specific model training or adaptation \cite{xie2024omnicontrol,pinyoanuntapong2025maskcontrol,watanabe2025simdiff,tashakori2025flexmotion,li2024controllable,zhang2024freemotion,cao2025motionctrl,li2025morph,hwang2025motionsynthesissparseflexible}, which, however, can generalize poorly to unseen constraint types \cite{liu2024programmable}. A few efforts have explored leveraging pre-trained motion models to generate constraint-aware motion in training-free manners \cite{liu2024programmable,karunratanakul2024optimizing}. These training-free methods depend on gradients computed from differentiable constraint objectives, yet real-world constraints can also be non-differentiable \cite{Louis_2024_BMVC,yuan2023physdiff}. This gap motivates us to develop a practically effective inference-time framework to integrate such heterogeneous constraints without requiring explicit differentiable formulations.

\textbf{Control Systems} \cite{naidu2018optimal,blaha2023survey} regulate behaviors of dynamical systems via control signals to achieve desired outcomes, and optimal control provides a theoretical framework for designing control laws under system and resource restrictions \cite{yong1999stochastic,theodorou2010generalized}. As diffusion models essentially define stochastic dynamical processes \cite{ho2020denoising,huang2024symbolic}, they can be viewed from a control perspective. 
Several recent works have explored leveraging control theory for tasks such as music generation~\cite{huang2024symbolic}, image generation~\cite{domingo-enrich2025adjoint,pandey2025variational,dou2026constrained}, video generation~\cite{Kim_2025_ICCV}, and phenotypic response simulation~\cite{Zhang_2026_CVPR}.
Different from these methods, we leverage the control-system view to specifically tackle the difficulty of jointly enforcing heterogeneous motion constraints with varying signal properties and spatial-temporal scopes, by developing a practically effective framework that incorporates and reconciles both criterion-based and objective-based constraints during motion generation.

\section{Proposed Method: MIC}
\label{sec:method}
\begin{figure}[t]
    \centering
    \includegraphics[width=0.9\linewidth]{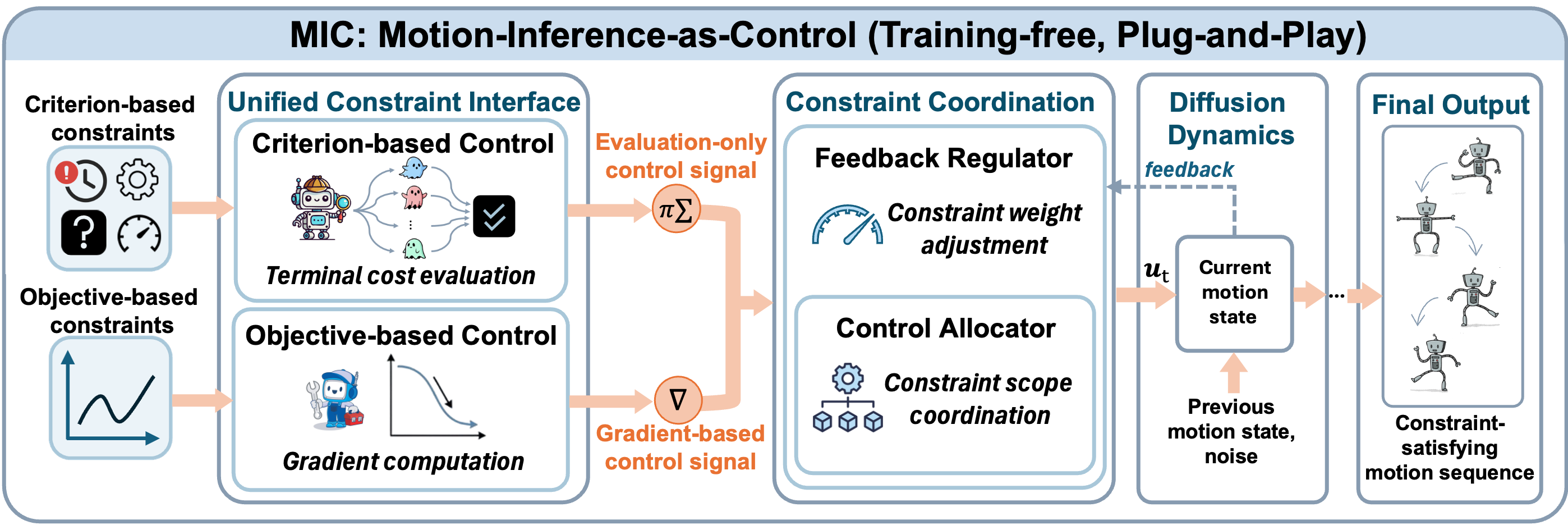}
    \caption{Illustration of our proposed MIC framework. MIC adopts a unified constraint interface that converts heterogeneous constraints into control signals: criterion-based constraints are handled via forward terminal costs, while continuous objective-based constraints employ gradient computation. A constraint coordination mechanism then integrates these signals, with a feedback regulator adapting constraint weights from violation levels and a control allocator coordinating constraint scopes across joints and frames. The resulting control $\mathbf{u}_t$ is injected at each diffusion step to steer denoising toward constraint satisfaction while minimally perturbing the pretrained motion prior.
    }
    \label{fig:framework}
\end{figure}

In this paper, we propose the MIC framework (as shown in Fig. \ref{fig:framework}), a training-free, plug-and-play constraint control framework that leverages pre-trained motion diffusion models to generate human motions that remain natural while satisfying diverse heterogeneous constraints, in a practically effective manner. 
To achieve this, diffusion-based controllable motion generation is reinterpreted as a controlled stochastic dynamical process. 
Based on this, a unified and practically effective step-wise guidance can be developed from optimal control theory: criterion-based constraints are enforced via evaluation feedback without requiring gradients, while continuous objective-based constraints are modeled via gradient-based control (Sec. \ref{sec:control_view}). Building on this shared control interface, MIC proposes a constraint coordination mechanism to adaptively balance and reconcile heterogeneous constraints over diffusion steps, reducing constraint imbalance and scope interference when composing multiple constraints (Sec. \ref{sec:coord_controller}).

\subsection{Background and Problem Setting}
\label{sec:background}
\textbf{Background on Diffusion-based Motion Generation Process.}
Given a text prompt, diffusion models \cite{tevet2023human,zhang2024motiondiffuse,meng2025rethinking,dabral2023mofusion} generate motion by progressively transforming Gaussian noise $\mathbf{x}_T \sim \mathcal{N}(\textbf{0}, \textbf{I})$ into a desired motion $\mathbf{x}_0$ via a series of denoising steps in a Markov chain. This is typically modeled as solving reverse-time stochastic differential equations (SDE) \cite{grenander1994representations,yuan2023physdiff,weng2026realign} from $t=T$ to $t=0$:
\begin{equation}
    \mathrm{d}\mathbf{x}_t = \big[-\frac{\beta(t)}{2}\mathbf{x}_t - \beta(t) \nabla_{\mathbf{x}_t}\log p_t(\mathbf{x}_t|c^p)\big]\mathrm{d}t +\sqrt{\beta(t)}\mathrm{d}\bar{\mathbf{\varepsilon}}_t,
    \label{eq:diffusion}
\end{equation}
where $\beta(t)$ is the noise schedule of the diffusion process, $\nabla_{\mathbf{x}_t}\log p_t(\mathbf{x}_t|c^p)$ denotes the (estimated) time-dependent score conditioned on a given text prompt $c^p$, $\mathrm{d}\bar{\mathbf{\varepsilon}}_t$ is a standard reverse-time Wiener process, and  $\mathrm{d}t$ represents negative infinitesimal timestep. 
Note that in Eq. \ref{eq:diffusion}, the widely used variance preserving (VP) form of SDE \cite{song2020score,chung2023diffusion,huang2024symbolic} is taken for concreteness, while it can be extended to other SDE variants. Also, the estimated score can be implemented under different denoising parameterizations \cite{ho2020denoising,Karunratanakul_2023_ICCV}, e.g., by predicting the clean motion state \cite{tevet2023human} or estimating the noise \cite{zhang2024motiondiffuse} (see Supplementary for more details).

\textbf{Problem Setting.} 
In constraint-aware motion generation~\cite{liu2024programmable,karunratanakul2024optimizing,Karunratanakul_2023_ICCV}, given a text prompt $c^p$ with a set of target constraints, the model is required to generate a text-conditioned motion sequence $\mathbf{x}\in\mathbb{R}^{N\times J\times D}$ that satisfies the target constraints, where the motion sequence contains $N$ frames of human poses $x_i\in \mathbb{R}^{J\times D}$, each represented by $D$-dimensional features (e.g., rotations or positions) of $J$ joints. 
We aim to leverage a pre-trained motion diffusion model as a text-conditioned motion prior $p(\mathbf{x}\,|\,c^p)$ and enforce the target constraints in a plug-and-play manner during inference.
For constraint target $\mathbf{d}$ and constraint evaluator $v_{\mathbf{d}}(\cdot)$, constraint satisfaction is modeled via an energy-based likelihood $p(\mathbf{d}|\mathbf{x})\propto \exp\big(-v_{\mathbf{d}}(\mathbf{x})\big)$ following \cite{huang2024symbolic,chung2023diffusion,zhang2025energyweighted,song2023loss}. Our goal is to sample motions from the following constraint-aware motion distribution:
\begin{equation}
    p(\mathbf{x}|c^p,\mathbf{d}) = \frac{p(\mathbf{x}|c^p) p(\mathbf{d}|\mathbf{x})}{A} \propto p(\mathbf{x}|c^p) \exp\big(-v_{\mathbf{d}}(\mathbf{x})\big),
    \label{eq:goal}
\end{equation}
where $A=\int_{\mathbf{x}} p(\mathbf{x}|c^p)p(\mathbf{d}|\mathbf{x})\mathrm{d}\mathbf{x}$ is a normalizing constant. Notably, the above distribution defines a target posterior that stays close to the motion prior $p(\mathbf x|c^p)$ (thereby maintaining motion naturalness) while biasing samples toward better constraint satisfaction.  
However, sampling from this posterior can be highly non-trivial in practice: constraint targets $\mathbf{d}$ are often heterogeneous and complex, and the evaluator $v_{\mathbf{d}}(\cdot)$ may provide discontinuous, sparse, or even black-box feedback, making direct constraint enforcement during sampling challenging.

\subsection{Optimal Control View of Controllable Motion Generation} \label{sec:control_view}
\textbf{Formulating controllable motion generation as optimal control.}
To tackle the challenges of enforcing heterogeneous constraints in motion generation, inspired by stochastic optimal control \cite{huang2024symbolic}, we cast inference-time controllable motion generation as an optimal control problem and further develop practically effective instantiations of the resulting control laws for both criterion-based and objective-based constraints. Below, we first show the connection between the motion generation dynamics and the stochastic control process.

To steer the sampling process (Eq. \ref{eq:diffusion}) of pre-trained motion diffusion model to instead sample from the desired distribution (Eq. \ref{eq:goal}), a step-wise guidance term $\mathbf{h}_t(\mathbf{x}_t,t)$ can be injected into the diffusion process, yielding: 
\begin{equation}
    \mathrm{d}\mathbf{x}_t=
    \Big[-\frac{1}{2}\beta(t)\mathbf{x}_t - \beta(t) \big(\nabla_{\mathbf{x}_t}\log p_t(\mathbf{x}_t\mid c^p) + \mathbf{h}_t(\mathbf{x}_t,t)\big)\Big]\mathrm{d}t + \sqrt{\beta(t)}\mathrm{d}\bar{\mathbf{\varepsilon}}_t.
    \label{eq:diff_guidance}
\end{equation}
The problem therefore becomes to find a proper $\mathbf{h}_t$ over time that guides the diffusion dynamics toward the desired posterior $p(\mathbf{x}|c^p, \mathbf{d})$, which is highly non-trivial when the constraints are criterion-based with sparse, non-smooth, or even black-box feedback. This motivates us to draw on the stochastic optimal control theory to characterize the desired step-wise guidance.

Specifically, Eq. \ref{eq:diff_guidance} can be interpreted from a control-system view \cite{yong1999stochastic,huang2024symbolic}. Intuitively, the diffusion sampling process can be viewed as a stochastic dynamical system evolving from the noisy state $\mathbf{x}_T$ to the final motion $\mathbf{x}_0$, with the step-wise guidance term $\mathbf h_t$ acting as control signals that steer the dynamics toward constraint satisfaction. To make this correspondence explicit, define the system state as $\mathbf{z}_t = \mathbf{x}_{T-t}$. The reverse diffusion process can be expressed as the following stochastic dynamical system (more details in Supplementary):
\begin{equation}
    \mathrm{d}\mathbf{z}_t=\mathbf{b}(\mathbf{z}_t,t)\mathrm{d}t +\sigma(\mathbf{z}_t,t)\big(\mathbf{u}_t(\mathbf{z}_t,t)\mathrm{d}t + \mathrm{d}\mathbf{\varepsilon}_t \big),
    \label{eq:control_system}
\end{equation}
where 
$\mathbf{b}(\mathbf{z}_t, t) = \frac{1}{2}\beta(T-t)\mathbf{z}_t + \beta(T-t)\nabla_{\mathbf{z}_t}\log p_{T-t}(\mathbf{z}_t\,|\, c^p)$, $\sigma(t) = \sqrt{\beta(T-t)}$, $\mathbf{u}_t(\mathbf{z}_t, t) = \sqrt{\beta(T-t)}\, \mathbf{h}_{T-t}(\mathbf{z}_t, T-t)$, $\mathrm{d}\mathbf{\varepsilon}_t$ is a standard Wiener process, $\mathrm{d}t$ is an infinitesimal timestep, and $t$ now denotes forward-time $0\rightarrow T$. We omit $\mathbf{z}_t$ in $\sigma(\mathbf{z}_t,t)$ and use $\sigma(t)$ in the following, as the noise scale depends only on $t$.

Then, building on this control process formulation, the problem of finding proper step-wise guidance in diffusion-based controllable motion generation can be cast as an optimal stochastic control problem \cite{huang2024symbolic}: finding a per-step \textit{control law} $\mathbf{u}^*_t(\mathbf{z}_t,t)$ that drives the stochastic dynamics (Eq. \ref{eq:control_system}) toward constraint satisfaction, while remaining close to the uncontrolled (prior) motion diffusion dynamics. The optimal control law $\mathbf{u}^*_t$ is obtained by minimizing the \textit{expected cumulative cost} $J_\mathbf{u}$ in the control system \cite{yong1999stochastic}. In our problem, $J_\mathbf{u}$ can be constructed as penalizing motion constraint violations while regularizing deviations from the natural-motion diffusion dynamics: 
\begin{equation}
    \mathbf{u}^*_t(\mathbf{z}_t, t)=\argmin_\mathbf{u} J_\mathbf{u}(\mathbf{z}_t, t) = \argmin_\mathbf{u} \mathbb{E} \big[\mathcal{E}(\mathbf{z}_T)+ \int_t^T L(\mathbf{z}_s,\mathbf{u}_s, s)  \mathrm{d}s \big], 
    \label{eq:cost_func}
\end{equation}
where $\mathcal{E}(\mathbf{z}_T)$ denotes the terminal cost induced by the motion constraint evaluator $v_{\mathbf d}$, penalizing constraint violations measured on the final generated motion $\mathbf z_T$, and $\int_t^T L(\mathbf{z}_s,\mathbf{u}_s, s) \mathrm{d}s=\int_t^T \frac{1}{2}\|\mathbf{u}_s(\mathbf{z}_s, s)\|^2 \mathrm{d}s$ is the running cost that penalizes control effort, encouraging minimal deviation from the motion prior.

\textbf{Solving the optimal control law.}
Under the above control formulation, the desired step-wise guidance can be characterized using established results from stochastic optimal control.
Specifically, applying the Hamilton--Jacobi--Bellman (HJB) equation \cite{yong1999stochastic} and Feynman--Kac formula \cite{oksendal2013stochastic,yong1999stochastic}, the optimal control law in Eq. \ref{eq:cost_func} takes the following analytic form~\cite{pavon1989stochastic,huang2024symbolic}:
\begin{equation}
\begin{aligned}
\mathbf{u}^*_t(\mathbf{z}_t,t)\,\mathrm{d}t
&= \sigma(t)\,\nabla_{\mathbf z_t}\log \psi(\mathbf{z}_t,t)\,\mathrm{d}t
\\
&= \sigma(t)\,\nabla_{\mathbf z_t}\log \mathbb{E}_{\mathcal P^0}\!\left[\exp\!\big(-\mathcal{E}(\mathbf{z}_T)\big)\ \middle|\ \mathbf{z}_t\right]\mathrm dt,
\end{aligned}
\label{eq:optimal_control1}
\end{equation}
where $\psi(\mathbf{z}_t,t) = \mathbb{E}_{\mathcal P^0}\!\left[\exp\!\big(-\mathcal{E}(\mathbf{z}_T)\big)\ |\ \mathbf{z}_t\right]$ denotes the desirability function associated with the optimal control law $\mathbf{u}_t^*$, and $\mathcal P^0$ represents the distribution over diffusion sampling trajectories under the uncontrolled diffusion dynamics, i.e., the text-conditioned natural motion prior dynamics with $\mathbf{u}_t\equiv 0$.

The resulting analytic form of the optimal control law $\mathbf{u}^*_t$ can serve as a principled guidance rule for constraint-aware motion generation. However, this still does not solve the above-mentioned first challenge, as applying the control in Eq. \ref{eq:optimal_control1} still requires gradient computation, which is unavailable when constraint feedback is provided only via evaluation signals (e.g., discontinuous rules or black-box simulators). This raises a practical question: \textit{can the optimal control law be expressed in a form that does not require gradient computation?}

\ul{Solving for criterion-based constraints.} 
Building on stochastic control theory \cite{theodorou2010generalized,huang2024symbolic}, the answer is affirmative. Specifically, the Markov structure of the motion diffusion process and the Gaussian structure of $\mathrm{d}\mathbf{\varepsilon}_t$ allow the tower rule \cite{williams1991probability} and Stein’s lemma \cite{stein1981estimation} to express the gradient of the desirability function as a conditional expectation over diffusion noise, yielding an equivalent control law that requires only forward constraint evaluations:
\begin{equation}
    \mathbf{u}^*_t(\mathbf{z}_t,t)\,\mathrm{d}t=\frac{\mathbb{E}_{\mathcal{P}^0}\big[\exp\big(-\mathcal{E}(\mathbf{z}_T)\big)\,\mathrm{d}\mathbf{\varepsilon}_t\big|\mathbf{z}_t\big]}{\mathbb{E}_{\mathcal{P}^0}\big[\exp\big(-\mathcal{E}(\mathbf{z}_T)\big)\big|\mathbf{z}_t\big]}
    \label{eq:optimal_expectation}
\end{equation}
Concretely, the resulting solution takes a path-integral form \cite{theodorou2010generalized} under the uncontrolled diffusion dynamics.
As Eq. \ref{eq:optimal_expectation} involves only forward evaluations of the constraint function, it naturally applies to the criterion-based motion constraints, regardless of whether the constraint feedback is intermittent, non-smooth, or only accessible through query-based evaluators. Moreover, \textit{Theorem 1} in the Supplementary shows that applying this control law in motion diffusion sampling recovers the target posterior in Eq. \ref{eq:goal}.

Having obtained this evaluation-based form for criterion-based constraints, we now turn to the next question: \textit{can we obtain an analytic form of the same control law for continuous objective-based constraints with accessible gradients?}

\ul{Solving for continuous objective-based constraints.}
From Eq. \ref{eq:optimal_control1}, the desirability function $\psi(\mathbf{z}_t,t)$ is defined as the expected constraint score of the final generated motion.
When the constraint objective is differentiable, this score is differentiable with respect to the current state, allowing the control law to be expressed in gradient form \cite{pavon1989stochastic,huang2024symbolic}:
\begin{equation}
    \mathbf{u}^*_t(\mathbf{z}_t,t)\,\mathrm{d}t=\sigma(t)\nabla_{\mathbf{z}_t} \log \psi(\mathbf z_t,t)\mathrm d t =\sigma(t)\nabla_{\mathbf{z}}\log p_t(\mathbf{d}|\mathbf z_t) \mathrm d t,
    \label{eq:optimal_gradient}
\end{equation}
where $\mathbf{d}$ is the target constraint conditions and $p_t(\mathbf d|\mathbf z_t)=\int p(\mathbf d|\mathbf z_T) p(\mathbf z_T|\mathbf z_t)\mathrm d \mathbf{z}_T$ represents the time-$t$ constraint likelihood obtained by marginalizing the terminal likelihood over the uncontrolled diffusion dynamics.

To this end, both criterion-based and continuous objective-based constraints can be represented through corresponding control laws that steer the diffusion sampling dynamics. As Eqs. \ref{eq:optimal_expectation} and \ref{eq:optimal_gradient} are equivalent characterizations of the same underlying optimal control law in Eq. \ref{eq:optimal_control1}, they enter the motion generation dynamics through the same step-wise control interface. This unified control formulation naturally accommodates heterogeneous constraints within a single inference-time framework, enabling principled joint enforcement of diverse constraint types.

\textbf{Instantiating control signals.} 
Having obtained the optimal control law, we next aim to instantiate it into practical step-wise control signals to apply to the motion diffusion process. Specifically, as both optimal control forms in Eq. \ref{eq:optimal_expectation} and Eq. \ref{eq:optimal_gradient} involve expectation or marginalization over the uncontrolled diffusion dynamics that are intractable in practice, we need to approximate each form. To derive practically effective instantiation for \textit{criterion-based constraints} where Eq. \ref{eq:optimal_expectation} takes the form of a weighted expectation over $\mathrm d\mathbf \varepsilon_t$, we construct the approximation via importance sampling. At each denoising step, we draw $M$ noise increments $\{\mathrm{d}\mathbf{\varepsilon}_t^{(m)}\}_{m=1}^M$ from a proposal distribution $q=\mathcal{N}(\mu,\Sigma)$, obtain the estimated final motion state $\hat{\mathbf{z}}_T^{(m)}$ via Tweedie's formula \cite{efron2011tweedie}, and evaluate the terminal cost $\mathcal{E}(\hat{\mathbf{z}}_T^{(m)})$. We then compute weights according to Eq. \ref{eq:optimal_expectation} and form the step-wise control as:
\begin{equation}
    \mathbf{u}_{t}(\mathbf{z}_t,t)\mathrm{d}t=\sum_{m=1}^M\pi_t^{(m)}\mathrm{d}\mathbf{\varepsilon}_t^{(m)}
    =\sum_{m=1}^M \frac{\tilde{\pi}_{t}^{(m)}}{\sum_{j=1}^M \tilde{\pi}_{t}^{(j)}}\,\mathrm{d}\mathbf{\varepsilon}_{t}^{(m)},
    \label{eq:sampling}
\end{equation}
where $\tilde{\pi}_t^{(m)}= \exp(-\mathcal{E}(\hat{\mathbf{z}}_T^{(m)})) \frac{\,p_0(\mathrm{d}\mathbf{\varepsilon}_t^{(m)})}{q(\mathrm{d}\mathbf{\varepsilon}_t^{(m)})}$, with $p_0$ representing the base diffusion noise distribution. To further improve the approximation, we adapt $q$ to reduce its mismatch with the target distribution $\exp(-\mathcal{E}(\hat{\mathbf{z}}_T))p_0$ via the cross-entropy method~\cite{botev2013cross}. Specifically, at each step, after drawing $M$ samples and evaluating their importance weights $\{\pi^{(m)}\}$, we select an elite subset of high-weight samples according to their importance weights and update $q$ using these elite samples~\cite{botev2013cross,kroese2007application}. This process can progressively concentrate the proposal on lower-cost regions, thereby increasing the likelihood of sampling noise increments that induce lower terminal costs and improving the practical effectiveness of the approximation.
More details are provided in Supplementary.

Meanwhile, for \textit{continuous objective-based constraints}, we instantiate Eq. \ref{eq:optimal_gradient} using a tractable approximation based on Tweedie's formula \cite{efron2011tweedie}. Specifically, at each denoising step, we use Tweedie's formula to obtain a denoised estimate $\hat{\mathbf z}_T$ of the clean motion from the current state $\mathbf z_t$, and evaluate the objective-defined gradient on $\hat{\mathbf z}_T$ to construct the step-wise control signal:
\begin{equation}
    \mathbf{u}_t(\mathbf{z}_t,t)\mathrm{d}t=\sigma(t)\nabla_{\mathbf{z}}\log p(\mathbf{d}|\hat{\mathbf{z}}_T)\mathrm dt.
    \label{eq:dps}
\end{equation}
The above approximation aligns with the gradient-based diffusion guidance formula \cite{chung2023diffusion}. 
To this point, we have instantiated practical control signals for both criterion-based and continuous objective-based constraints, which can be injected into the motion generation dynamics through the same mechanism, thereby addressing the first challenge. We next turn to the second challenge: \textit{how to coordinate multiple heterogeneous constraints during the dynamic sampling process.}

\subsection{Constraint Coordination Mechanism} \label{sec:coord_controller}
In this mechanism, we tackle the challenge where multiple heterogeneous constraints are simultaneously required in generating a certain motion sequence. Specifically, given $K$ motion constraints for a motion sequence, based on Eqs. \ref{eq:sampling} and \ref{eq:dps}, we can derive $K$ corresponding control signals $\{\mathbf u_{t,k}\}_{k=1}^K$. A straightforward approach to combine them into the final guidance is via a weighted sum $\sum_{k=1}^K W_k \mathbf u_{t,k}$, where $W_k$ represents the weight for the $k$-th constraint. Yet, this alone may not be sufficient for reliable motion generation. In practical tasks, heterogeneous constraints often co-exist, with feedback that varies in signal strength and effective scope over the motion sequence (e.g., a single-frame position target versus a full-sequence constraint). Naively summing these control signals may induce constraint imbalance and interference, degrading motion coherence.

This motivates us to introduce a dedicated coordination mechanism on top of the per-constraint guidance signals. Notably, as we cast all constraint feedback as control signals under a shared control interface, coordinating heterogeneous constraints can be viewed as a structured control-design problem along two complementary axes: how strongly each constraint should act over diffusion steps, and where its effects should be allocated across joints and frames. Inspired by control system design \cite{aastrom2021feedback,blaha2023survey}, we propose a \textit{constraint coordination mechanism} with two components to stabilize multi-constraint guidance during motion generation: a feedback regulator that adapts constraint strengths, and a control allocator that distributes control effort across joints and frames.

\textbf{Feedback regulator.}
To handle constraint imbalance, rather than fixing or manually tuning constraint weights, we treat the weights as closed-loop control variables, as the appropriate weight for a constraint depends on how strongly it is violated, which evolves throughout the denoising process. We therefore draw on the principle of integral feedback control \cite{aastrom2021feedback}: if a constraint remains violated, its weight should accumulate over time, increasing pressure until the violation is resolved. Specifically, let $\tilde{c}_{k,t}\ge 0$ denote the violation level of the $k$-th constraint at step $t$, computed from its evaluator $v_{\mathbf d_k}(\cdot)$ on the predicted clean motion at the current step, i.e., $\tilde{c}_{k,t}=\max\big(0,\ v_{\mathbf d_k}(\hat{\mathbf z}_T)\big)$. We normalize $\tilde{c}_{k,t}$ by an exponential moving average-based running scale to obtain $c_{k,t}$, so that violations from heterogeneous constraints are comparable. The feedback regulator then updates $W_{k,t+1}$ (i.e., the weight of the $k$-th constraint at the $t+1$-th timestep) as:
\begin{equation}
\setlength{\abovedisplayskip}{3pt}
\setlength{\belowdisplayskip}{3pt}
    W_{k,t+1}=\Pi_{[0,W_{\max}]}(W_{k,t}+\gamma\,c_{k,t}),
    \label{eq:i_controller}
\end{equation}
where $\gamma$ is the integral gain and $\Pi_{[0,W_{\max}]}$ projects the weight onto a valid range. This integral-style weight adjustment rule has two properties: (1) if a constraint remains violated across steps, its weight increases, applying growing pressure until the violation is resolved; (2) once a constraint becomes satisfied ($\tilde{c}_{k,t} = 0$), its weight stops accumulating and remains stable, preventing it from unnecessarily dominating the control budget.
We provide more details in Supplementary.

\textbf{Control allocator.}
As each constraint can operate on different scopes of joints and frames, we view the joint-frame blocks in the motion as (spatial-temporal) control channels. This allows constraint composition to be treated as a \textit{budget allocation} problem \cite{blaha2023survey} rather than simply aggregating signals. In particular, each constraint proposes a desired control (i.e., $\mathbf{u}_{k,t}$), but it is only trusted on its effective spatial-temporal scope $M_k$. The allocator then seeks a single applied control $\mathbf{u}_t$ that best agrees with all constraints {on their respective scopes} while keeping the total control effort small, thus encouraging minimal deviation from the pre-trained motion prior. When conflicting constraint control signals are proposed, this allocation step yields a single applied control consensus that reduces cross-scope interference.
At step $t$, the applied control $\mathbf{u}_t$ is computed by solving a weighted least-squares allocation problem: 
\begin{equation}
    \mathbf{u}_t = \argmin_\mathbf{u} \sum_{k=1}^{K} \|W_{k,t}(M_k \mathbf{u} - M_k \mathbf{u}_{k,t})\|^2 + \lambda \|\mathbf{u}\|^2,
    \label{eq:allocation}
\end{equation}
where $M_k$ represents the control scope relevant to the $k$-th constraint, $W_{k,t}$ encodes its weight at step $t$ (obtained via Eq. \ref{eq:i_controller}), and $\lambda$ is a positive scalar that penalizes the total control effort, preventing excessive deviation from the pre-trained motion prior. 
Note that for notational convenience, in the above Eq.~\ref{eq:allocation}, we flatten the control signal into $\mathbf{u} \in \mathbb{R}^{NJD}$, and accordingly, $M_k \in \mathbb{R}^{NJD \times NJD}$ is a diagonal mask matrix that projects the control signal onto the effective scope of the $k$-th constraint. This weighted least-squares problem is quadratic and admits a closed-form solution:
\begin{equation}
\mathbf{u}_t = \Big(\sum_{k=1}^K W_{k,t}^2 M_k^\top M_k +\lambda I \Big)^{-1} \Big( \sum_{k=1}^K W_{k,t}^2 M_k^\top M_k \mathbf{u}_{k,t} \Big),
\end{equation}
which can be efficiently computed at each denoising step. 

Taken together, by jointly regulating \textit{how strongly} and \textit{where} each constraint influences the motion generation dynamics, MIC enables stable multi-constraint composition while preserving the natural motion prior.

\subsection{Overall Inference}
Our MIC framework operates in a fully training-free manner. As shown in Fig. \ref{fig:framework}, given a text prompt $c^p$ and $K$ motion constraints, MIC generates constraint-satisfying and natural motions by iteratively guiding the reverse diffusion sampling process. Specifically, at each denoising step $t$, MIC (i) computes the control signal $\mathbf{u}_{k,t}$ for each constraint, using Eq. \ref{eq:sampling} for criterion-based constraints and Eq. \ref{eq:dps} for continuous objective-based constraints; (ii) updates the constraint weights via the feedback regulator in Eq. \ref{eq:i_controller}; and (iii) resolves interference among constraints via control allocator in Eq. \ref{eq:allocation} to obtain the applied control $\mathbf{u}_t$, which is injected into the diffusion dynamics to produce the next state. An overall algorithm is provided in Supplementary.

\section{Experiments}
\label{sec:experiments}
\textbf{Benchmark.} To evaluate the effectiveness of the proposed MIC framework, we conduct experiments on the very comprehensive training-free controllable motion generation benchmark \cite{liu2024programmable}, including two different scenarios: \textit{open-set} (Tab.~\ref{tab:1}, with control constraints not sourced from motion datasets to evaluate open-set control capability) and \textit{known} (Tab.~\ref{tab:2}, with control constraints sampled from motion dataset \cite{zhang2023generating}). Following the benchmark, generated motions are evaluated on a series of requirements (constraints) containing both continuous objective-based and criterion-based ones. The control tasks include human-scene interaction (HSI), geometric constraints (GEO), and human-object interaction (HOI), with HSI-1 for key-frame head-height control, HSI-2 for barrier avoidance, HSI-3 for limited-area walking, GEO-1 for wall touching, and HOI-1 for object moving.
Meanwhile, the generated human motions are also required to remain realistic \cite{liu2024programmable,Karunratanakul_2023_ICCV}, such as maintaining temporal coherence and avoiding foot skating. We also evaluate whether the generated motions can pass a physical simulation check \cite{mujoco}.
More details are in the Supplementary.

\textbf{Metrics.}
Following \cite{liu2024programmable}, we evaluate motion quality using maximum joint acceleration (\textit{Max Acc.}) for frame-wise consistency and foot skating ratio (\textit{Skating}) for motion coherence. We also report the pass rate (\textit{Pass}) of motions that pass physical simulation checks in MuJoCo \cite{mujoco}. For per-task constraint satisfaction, we report constraint error (\textit{C.Err}) and unsuccess rate (\textit{Unsucc. Rate}, percentage of motions failing constraints) following \cite{liu2024programmable}. Additionally, for the \textit{known} scenario (HSI-1 in Tab. \ref{tab:2}) where constraint heights are sampled from the HumanML3D test set \cite{zhang2023generating}, we evaluate generative quality using FID, Diversity, and R-Precision \cite{shafir2023human} following \cite{liu2024programmable}. Details are provided in Supplementary.

\textbf{Implementation Details.}
We follow the benchmark implementation in \cite{liu2024programmable} to adopt the official Motion Diffusion Model (MDM) pre-trained on HumanML3D dataset \cite{zhang2023generating} and use its DDIM version for fair comparison. Note that our framework can also be applied to other diffusion models.
We set the sampling size in Eq. \ref{eq:sampling} as $M=16$.
The proposal distribution $q = \mathcal{N}(\boldsymbol{\mu}, \boldsymbol{\Sigma})$ is initialized as a standard Gaussian and updated at each denoising step using the cross-entropy method~\cite{botev2013cross} with elite ratio 20\%. 
We follow the benchmark implementation to construct the continuous objective-based constraints (e.g., head height constraints) and apply Tweedie's formula~\cite{efron2011tweedie} to obtain the denoised estimate $\hat{\mathbf{z}}_T$ at each step. For criterion-based constraints, such as rule-based criteria (e.g., foot-skating and success checks), we evaluate the corresponding constraint criteria to guide the motion generation for our method. For a stable initialization, we adopt the optimization strategy used in prior methods~\cite{karunratanakul2024optimizing,liu2024programmable} as a warm start, and subsequently apply MIC for step-wise constraint control and coordination. More details about the parameters and implementation are in Supplementary.

\subsection{Experiment Results}

\begin{table*}[t]
    \centering
    \caption{Comparison with other methods on various open-set constraint specifications following \cite{liu2024programmable}. MDM (Unconstrained) serves as an uncontrollable baseline. All methods are implemented based on the same pre-trained MDM. }
    \resizebox{0.85\linewidth}{!}{
    \begin{tabular}{l|*{5}{M{\metricw}}|*{5}{M{\metricw}}} 
    \toprule
    & \multicolumn{5}{c|}{Task HSI-2} &  \multicolumn{5}{c}{Task HSI-3}  \\
    \midrule
   Method & Skating$\downarrow$  & Max Acc.$\downarrow$ &  C.Err.$\downarrow$ & Unsucc. Rate$\downarrow$ & Pass$\uparrow$ & Skating$\downarrow$   & Max Acc.$\downarrow$ &  C.Err.$\downarrow$ &Unsucc. Rate$\downarrow$  & Pass$\uparrow$\\  
   \midrule
   MDM (Unconstrained) \cite{tevet2023human} & 0.096 & 0.126 & 0.454 & 1.000 & 0.000 &  0.096 & 0.126 & 0.301  & 0.875 & 0.125\\ \midrule
   IK \cite{liu2024programmable}             & 0.253 & 0.923 & 0.047  & 0.531 & 0.500 &  0.139  & 0.292 & 0.015  & 0.406 & 0.531 \\
   IK+Reg. \cite{liu2024programmable}        & 0.478 & 0.352 & 0.047 & 0.531 & 0.469 & 0.215 & 0.128 & 0.015  & 0.406 & 0.531 \\  
   DNO \cite{karunratanakul2024optimizing} &0.196 & 0.162 & 0.051 & 0.375 & 0.531 & 0.144 & 0.094 & 0.014 & 0.344 & 0.594 \\ 
   ProgMoGen \cite{liu2024programmable}   & 0.180 & 0.150 & 0.097 & 0.219 & 0.563 & 0.125 & 0.093 & 0.012  & 0.344 & 0.594\\ 
   ReAlign \cite{weng2026realign} & 0.245 & 0.155 & 0.067 & 0.250 & 0.469 & 0.133 & 0.102 & 0.016 & 0.594  & 0.375 \\
   \midrule
   Ours & \textit{0.172} & \textit{0.140} & \textit{0.009} & \textit{0.094} & \textit{0.875} & \textit{0.112} & \textit{0.089} & \textit{0.004} & \textit{0.250} & \textit{0.719} \\ 
  \bottomrule
  \toprule
    & \multicolumn{5}{c|}{Task GEO-1} &  \multicolumn{5}{c}{Task HOI-1}  \\
    \midrule
   Method & Skating$\downarrow$  & Max Acc.$\downarrow$ &  C.Err.$\downarrow$ & Unsucc. Rate$\downarrow$ & Pass$\uparrow$ & Skating$\downarrow$ & Max Acc.$\downarrow$ &  C.Err.$\downarrow$ & Unsucc. Rate$\downarrow$ & Pass$\uparrow$ \\  
   \midrule
   MDM (Unconstrained) \cite{tevet2023human}   & 0.096 & 0.126 & 0.233 & 1.000 & 0.000& 0.029  & 0.026 & 1.701   & 1.000 & 0.000\\ \midrule
   MDM Edit \cite{tevet2023human}         & 0.161 & 0.147 & 0.141 & 1.000 & 0.000 & 0.029 & 0.032 & 1.739  & 1.000 & 0.000 \\
   PriorMDM \cite{shafir2023human}         & 0.350 & 0.197 & 0.185 & 1.000 & 0.000 & 0.327 & 0.213 & 1.884   & 1.000 & 0.000 \\
   IK \cite{liu2024programmable}                & 0.147 & 0.187 & 0.010 & 0.219 & 0.625 & 0.408 & 0.919 & 0.011  & 0.125 & 0.875 \\
   IK+Reg. \cite{liu2024programmable}          & 0.536 & 0.117 & 0.010 & 0.219 & 0.625 & 0.267 & 0.153 & 0.011  & 0.125 & 0.875 \\ 
   DNO \cite{karunratanakul2024optimizing} & 0.116  & 0.113 & 0.035 & 0.563 & 0.375 & 0.110 & 0.086 & 0.035 & 0.219  & 0.719 \\ 
   ProgMoGen \cite{liu2024programmable}  & 0.110 & 0.104 & 0.023 & 0.531 & 0.406 & 0.109 & 0.067 & 0.028  & 0.188  & 0.750\\ 
   ReAlign \cite{weng2026realign} & 0.122 & 0.132 & 0.056 & 0.625 & 0.313 & 0.193 & 0.169 & 0.033 & 0.250 & 0.625 \\
    \midrule
   Ours & \textit{0.102} & \textit{0.088} & \textit{0.008} & \textit{0.094} & \textit{0.781} & \textit{0.036} & \textit{0.062} & \textit{0.004} & \textit{0.031}  & \textit{0.938}\\ 
  \bottomrule
\end{tabular}}
\label{tab:1}
\end{table*}

\begin{table}[t]
    \centering
    \caption{Comparison with other methods with constraints sampled from ground-truth HumanML3D test set following \cite{liu2024programmable}. MDM (Unconstrained) serves as an uncontrollable baseline. All methods are implemented based on the same pre-trained MDM.}
    \resizebox{0.75\linewidth}{!}{
    \begin{tabular}{l|*{5}{M{\metricw}}|*{3}{M{\metricw}}} 
    \toprule
     \multicolumn{9}{c}{Task HSI-1} \\
    \toprule
   Method & Skating$\downarrow$   & Max Acc.$\downarrow$ &  C.Err.$\downarrow$ & Unsucc. Rate$\downarrow$ & Pass$\uparrow$  & FID$\downarrow$ &  Diversity$\rightarrow$ & R-prec. (Top3)$\uparrow$\\  
   \midrule
   MDM (Unconstrained) \cite{tevet2023human}    & 0.086 & 0.097 & 0.118 & 0.718 & 0.193 & 0.545 & 9.656 & 0.610 \\ \midrule
   MDM Edit \cite{tevet2023human} & 0.094 & 0.148 & 0.109 & 0.645 & 0.303 & 0.554 & 9.656 & 0.614  \\
   IK  \cite{liu2024programmable} & 0.093 & 0.414 & 0.012 & 0.088 & 0.735 & 0.545 & 9.653  & 0.610\\
   IK+Reg.  \cite{liu2024programmable} & 0.269 & 0.121 & 0.012 & 0.088 & 0.739 & 0.782 & 9.509  & 0.603\\   
   DNO \cite{karunratanakul2024optimizing} & 0.085 & 0.098 & 0.018 & 0.121 & 0.765 & 0.559 & 9.595 & 0.588 \\
   ProgMoGen \cite{liu2024programmable}    & 0.075 & 0.094 & 0.012 & 0.088 & 0.776 & 0.556 & 9.611   & 0.597 \\ 
   ReAlign \cite{weng2026realign} & 0.096 & 0.104 & 0.024 & 0.189 & 0.757 & 0.619 & 9.433 & 0.635 \\
   \midrule
   Ours & \textit{0.074} & \textit{0.093} & \textit{0.009} & \textit{0.068} & \textit{0.857} & \textit{0.494} & \textit{9.656} & \textit{0.635} \\
  \bottomrule
    \end{tabular}}
    \label{tab:2}
\end{table}

\subsubsection{Quantitative Results.}
We compare against previous gradient-based methods \cite{liu2024programmable,karunratanakul2024optimizing,weng2026realign,shafir2023human}, all using the same MDM backbone. As these methods require differentiable objectives, we use surrogate proxies for criterion-based constraints (e.g., replacing discrete success checks with continuous distance-based scores). As shown in Tab. \ref{tab:1}--\ref{tab:2}, MIC achieves strong overall performance across all tasks. Notably, we observe substantial improvements on criterion-based metrics such as unsuccess rate and simulation check pass rate, while achieving high motion quality. This demonstrates MIC's effectiveness in handling heterogeneous constraints while maintaining motion generation quality. We provide more details in Supplementary.

\begin{wrapfigure}{r}{0.5\linewidth}
    \centering
    \includegraphics[width=\linewidth]{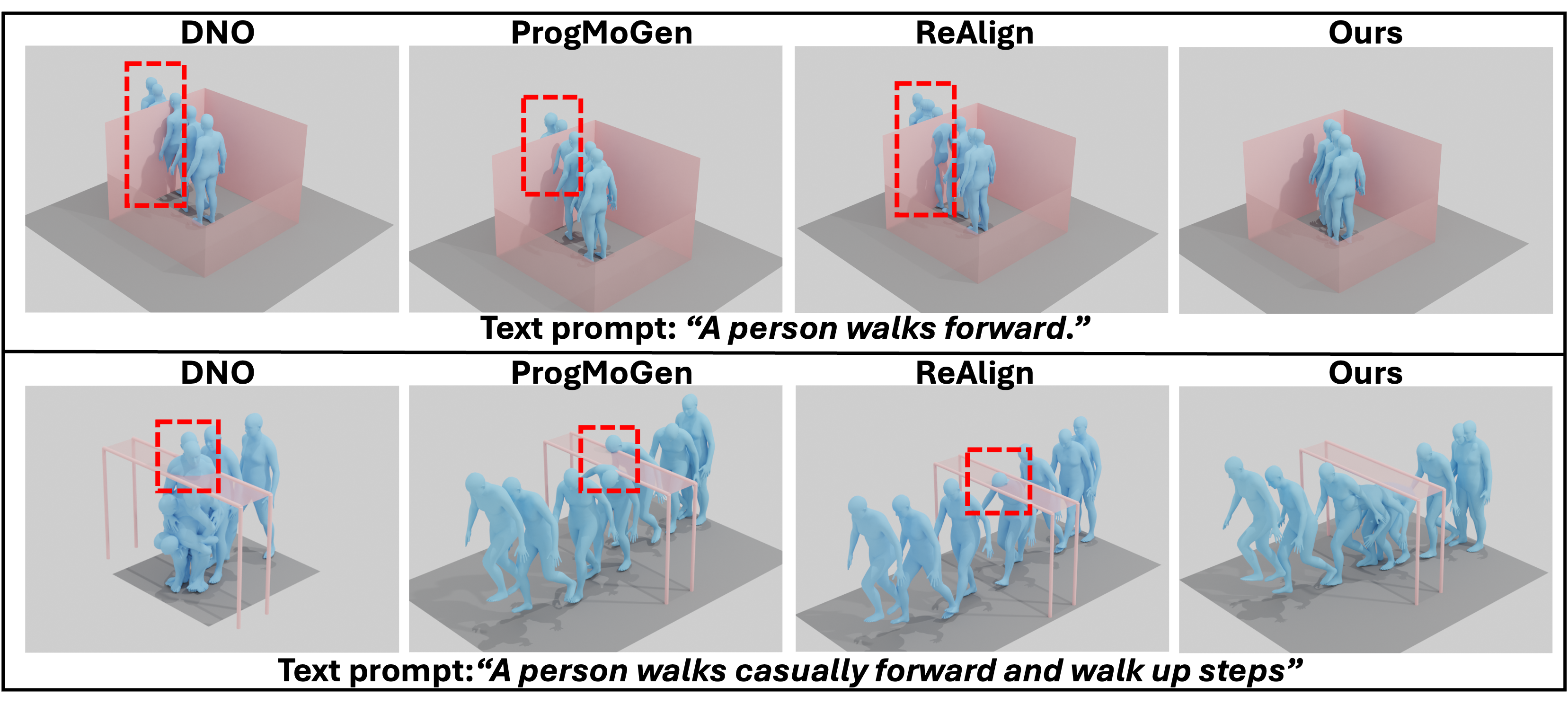}
    \caption{Visualization comparisons to previous methods. The compared methods produce motions with constraint violations such as boundary penetration and collisions (highlighted in red boxes), whereas our method consistently satisfies constraints while maintaining natural motion quality.}
    \label{fig:qualitative}
\end{wrapfigure}
\subsubsection{Qualitative Results.}
\begin{figure}[t]
    \centering
    \includegraphics[width=0.95\linewidth]{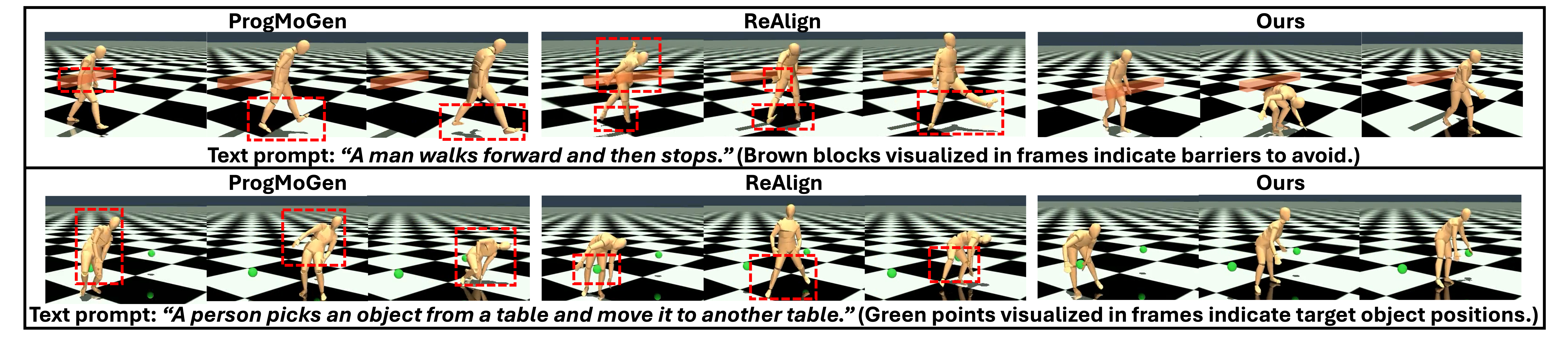}
    \caption{Qualitative comparisons with ProgMoGen \cite{liu2024programmable} and ReAlign \cite{weng2026realign}
    in physical simulation. The motions generated by the compared methods exhibit clear constraint violations and unstable movements (highlighted in red boxes), while MIC consistently satisfies different constraints while maintaining motion naturalness.}
    \label{fig:qualitative_compare}
\end{figure}

We also show qualitative comparisons. As shown in \cref{fig:qualitative}, our method produces motions that better satisfy constraints while maintaining motion naturalness and coherence. Moreover, we also incorporate physics-based evaluation \cite{mujoco,luo2022embodied} to serve as an external constraint evaluator for criteria such as maintaining motion stability. As shown in \cref{fig:qualitative_compare}, while the compared methods often produce unstable motions or fail to satisfy control requirements, our method can generate motions that successfully meet the corresponding  requirements. 

\subsection{Ablation Study}
Below we conduct ablation studies on HSI-2 task. \textbf{More ablation studies, further analysis, and user studies are provided in Supplementary.}

\begin{wraptable}{10}{0.45\linewidth}
\scriptsize
\caption{Impact of criterion-based constraint formulation.}
\centering
\resizebox{\linewidth}{!}{
\begin{tabular}{l*{5}{M{\metricw}}} 
    \toprule
   Method            & Foot Skate$\downarrow$  & Max Acc.$\downarrow$  & C.Err.$\downarrow$ & Unsucc. Rate$\downarrow$ & Pass$\uparrow$ \\  
   \midrule
   Baseline A    & 0.180 & 0.152 & 0.097 & 0.219 & 0.594  \\   
   Baseline B     & 0.242 & 0.267 & 0.205 & 0.281 & 0.594 \\ 
   Baseline C     & 0.274 & 0.243 & 0.231 & 0.250 & 0.625 \\ 
   Baseline D     & 0.208 & 0.238 & 0.147 & 0.219 & 0.625 \\ 
    \midrule
   \textit{MIC}  & 0.172 & 0.140 & 0.009 & 0.094 & 0.875  \\   
  \bottomrule
    \end{tabular}}
    \label{tab:abl_criterion}
\end{wraptable}
\textbf{Impact of criterion-based constraint formulation.} 
To evaluate our framework for handling criterion-based constraints, we compare against the following four variants.
\textbf{Baseline A} replaces criterion-based constraints with surrogates and uses gradients to guide diffusion sampling. \textbf{Baseline B} estimates gradients of the non-differentiable criteria via zeroth-order optimization \cite{spall2002multivariate}. \textbf{Baseline C} applies reinforcement learning \cite{williams1992simple} to approximate gradient guidance for criterion-based constraints. \textbf{Baseline D} employs evolutionary strategies \cite{salimans2017evolution} to search for effective guidance signals. As shown in \cref{tab:abl_criterion}, our MIC outperforms all baselines, with a significant improvement on criterion-based constraints such as unsuccess rate. This may be because surrogate proxies can be misaligned with the true criteria, while gradient estimation can be unstable with complex constraints under diffusion dynamics. In contrast, MIC directly constructs control signals for criterion-based constraints without requiring gradients, providing effective guidance.

\begin{wraptable}{r}{0.55\linewidth}
\scriptsize
\caption{Impact of unified mechanism.}
\centering
\resizebox{\linewidth}{!}{
\begin{tabular}{l*{5}{M{\metricw}}} 
    \toprule
   Method            & Foot Skate$\downarrow$  & Max Acc.$\downarrow$  & C.Err.$\downarrow$ & Unsucc. Rate$\downarrow$ & Pass$\uparrow$ \\  
   \midrule
   Separate handling    & 0.247 & 0.165 & 0.085 & 0.188 & 0.656  \\   
    \midrule
   \textit{Unified handling (MIC)}  & 0.172 & 0.140 & 0.009 & 0.094 & 0.875  \\   
  \bottomrule
    \end{tabular}}
    \label{tab:abl_joint}
\end{wraptable}
\textbf{Impact of unified framework for heterogeneous constraints.}
Our framework handles criterion-based constraints and continuous objective-based constraints within the same inference time mechanism. To evaluate the benefit of such integration, we compare with the \textbf{Separate handling} variant that uses other gradient-based mechanisms in previous methods for continuous objective-based constraints and our mechanism for criterion-based constraints and report the best results. As shown in Tab. \ref{tab:abl_joint}, our unified mechanism achieves better performance, demonstrating its effectiveness.

\begin{wraptable}{r}{0.47\linewidth}
\scriptsize
\caption{Impact of the constraint coordination mechanism.}
\centering
\resizebox{\linewidth}{!}{
\begin{tabular}{l*{5}{M{\metricw}}} 
    \toprule
   Method            & Foot Skate$\downarrow$  & Max Acc.$\downarrow$  & C.Err.$\downarrow$ & Unsucc. Rate$\downarrow$ & Pass$\uparrow$ \\  
   \midrule
    w/o regulation     & 0.208 & 0.143 & 0.014 & 0.125 & 0.813 \\ 
    w/o allocation    & 0.189 & 0.142 & 0.048 & 0.125 & 0.750  \\   
    w/o coordination & 0.220 & 0.146 & 0.054 & 0.156 & 0.688 \\ 
    \midrule
   \textit{MIC}  & 0.172 & 0.140 & 0.009 & 0.094 & 0.875  \\   
  \bottomrule
    \end{tabular}}
    \label{tab:abl_loop}
\end{wraptable}
\textbf{Impact of the constraint coordination mechanism.}
In MIC, we propose a constraint coordination mechanism with feedback regulator and control allocator to adaptively coordinate heterogeneous constraints during motion generation. To evaluate this, we conduct experiments with the following variants: \textbf{w/o regulation} that only performs control allocation (Eq. \ref{eq:allocation}) without adaptively adjusting the weights of different constraints using feedback regulation; \textbf{w/o allocation} that only adjusts the constraint weights via the feedback regulator (Eq. \ref{eq:i_controller}) without performing control allocation; and \textbf{w/o coordination} that directly averages different control signals without feedback regulation or control allocation. As shown in \cref{tab:abl_loop}, MIC achieves the best result, demonstrating the efficacy of our design.

\begin{wrapfigure}{r}{0.5\linewidth}
    \centering
    \includegraphics[width=\linewidth]{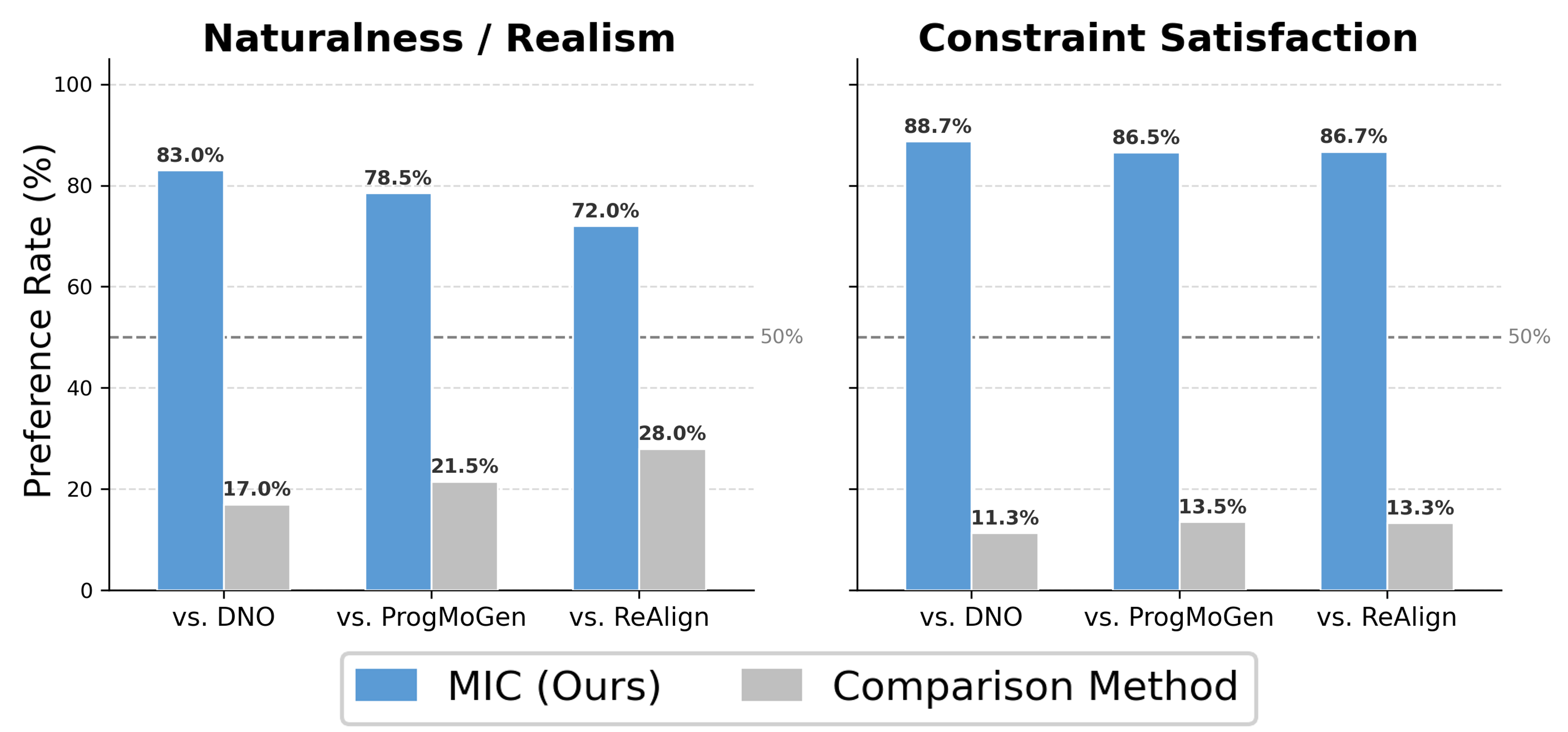}
    \caption{User study. We report the preference rate (\%) of pairwise comparisons between our MIC and each compared method. }
    \label{fig:user_study}
\end{wrapfigure}
\textbf{User study.}
Following \cite{chen2023executing,tevet2023human}, we conduct a user study to assess motion quality. We compare MIC with DNO~\cite{karunratanakul2024optimizing}, ProgMoGen~\cite{liu2024programmable}, and ReAlign~\cite{weng2026realign}, using 30 motions generated from the same prompts and constraint settings. Twenty participants complete paired comparisons between MIC and each baseline, judging (i) motion naturalness and realism, and (ii) constraint satisfaction, with randomized left-right ordering. Fig. \ref{fig:user_study} reports the preference rates for MIC, which is consistently favored on both criteria.

\section{Conclusion}
In this paper, we propose MIC, a training-free framework for guiding pre-trained motion diffusion models to generate natural, constraint-satisfying human motion under heterogeneous real-world constraints. We design a novel inference-time control framework that injects heterogeneous signals into the sampling dynamics via a constraint coordination mechanism that stabilizes and coordinates the constraints. Experiments on tasks with diverse constraint requirements demonstrate the effectiveness of our framework.

%
%
\bibliographystyle{splncs04}
\bibliography{main}
\end{document}